\documentclass[10pt,twocolumn,letterpaper]{article}

\usepackage{cvpr}
\usepackage{times}
\usepackage{epsfig}
\usepackage{graphicx}
\usepackage{amsmath}
\usepackage{amssymb}
\usepackage{multirow}
\usepackage{booktabs}

\usepackage[breaklinks=true,bookmarks=false]{hyperref}

\cvprfinalcopy 

\def\cvprPaperID{****} 

\begin{document}

\title{DM-GAN: Dynamic Memory Generative Adversarial Networks for Text-to-Image Synthesis}

\author{
Minfeng Zhu$^{1,3}$\thanks{This work was done when Minfeng Zhu was visiting the University of Technology Sydney.}\hspace{2em}Pingbo Pan$^3$\hspace{2em}Wei Chen$^1$\hspace{2em}Yi Yang$^{2,3}$\thanks{Part of this work was done when Yi Yang was visiting Baidu Research during his Professional Experience Program.}\\
$^1$ State Key Lab of CAD\&CG, Zhejiang University\hspace{2em}$^2$ Baidu Research\\
$^3$ Centre for Artificial Intelligence, University of Technology Sydney \\
{\tt\small \{minfeng\_zhu@, chenwei@cad\}zju.edu.cn\hspace{2em}\{pingbo.pan@student,Yi.Yang@\}uts.edu.au}
}

\maketitle

\begin{abstract}
In this paper, we focus on generating realistic images from text descriptions. Current methods first generate an initial image with rough shape and color, and then refine the initial image to a high-resolution one. Most existing text-to-image synthesis methods have two main problems. (1) These methods depend heavily on the quality of the initial images. If the initial image is not well initialized, the following processes can hardly refine the image to a satisfactory quality. (2) Each word contributes a different level of importance when depicting different image contents, however, unchanged text representation is used in existing image refinement processes. In this paper, we propose the Dynamic Memory Generative Adversarial Network (DM-GAN) to generate high-quality images. The proposed method introduces a dynamic memory module to refine fuzzy image contents, when the initial images are not well generated. A memory writing gate is designed to select the important text information based on the initial image content, which enables our method to accurately generate images from the text description. We also utilize a response gate to adaptively fuse the information read from the memories and the image features. We evaluate the DM-GAN model on the Caltech-UCSD Birds 200 dataset and the Microsoft Common Objects in Context dataset. Experimental results demonstrate that our DM-GAN model performs favorably against the state-of-the-art approaches.
\end{abstract}
\section{Introduction}
\begin{figure}[!t]
    \centering
    \includegraphics[width=0.99\columnwidth]{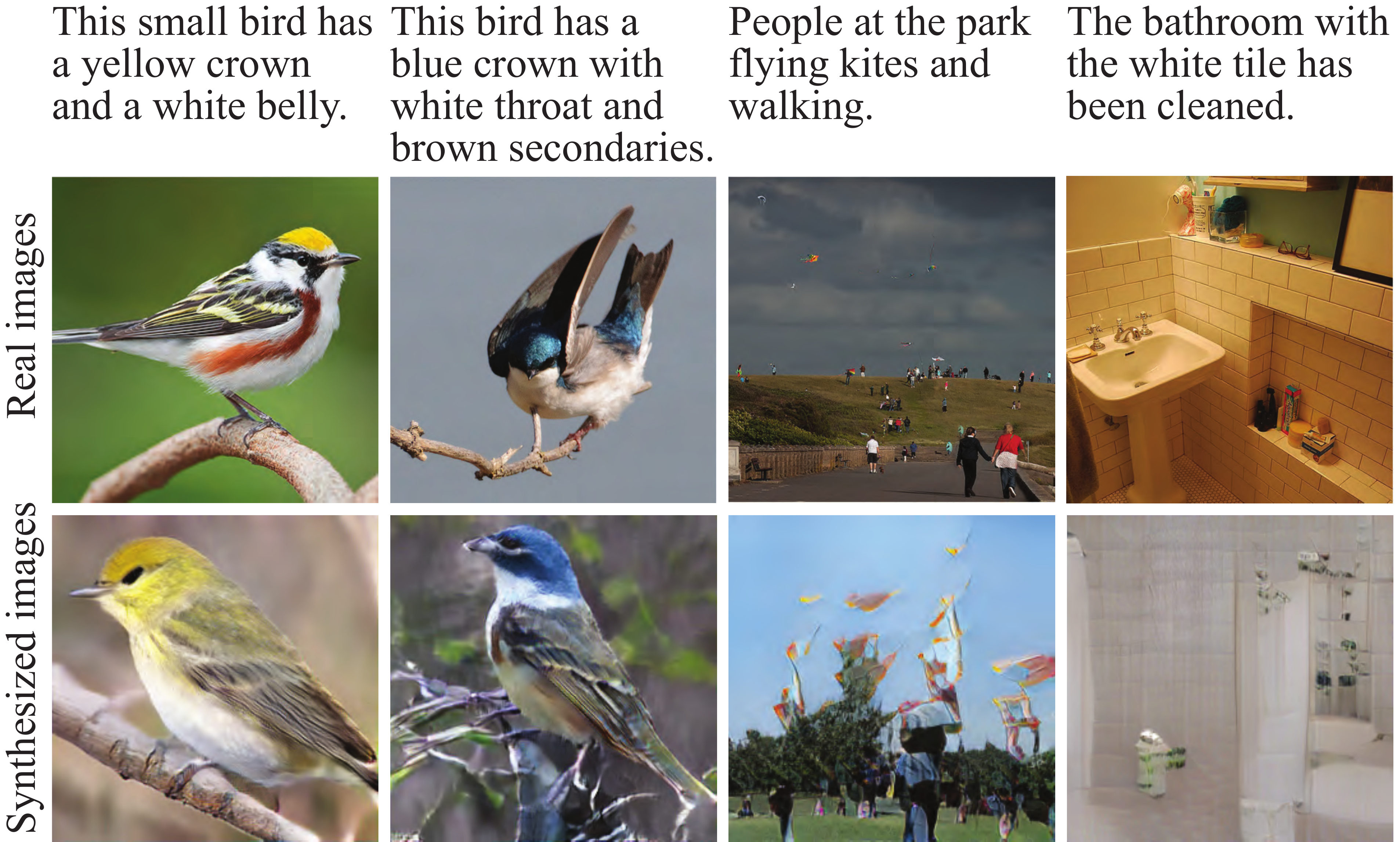}
    \caption{Examples of text-to-image synthesis by our DM-GAN.}
    \label{fig:example}
\end{figure}

The last few years have seen remarkable growth in the use of Generative Adversarial Networks (GANs) \cite{Goodfellow:2014td} for image and video generation.
Recently, GANs have been widely used to generate photo-realistic images according to text descriptions (see Figure \ref{fig:example}).
Fully understanding the relationship between visual contents and natural languages is an essential step towards artificial intelligence, \eg, image search and video understanding \cite{Zhu_2017_CVPR}.
Multi-stage methods \cite{Xu:2017wg,Zhang:2017bw,zhang2017stackgan++} first generate low-resolution initial images and then refine the initial images to high-resolution ones.

Although these multi-stage methods achieve remarkable progress, there remain two problems.
First, the generation result depends heavily on the quality of initial images.
The image refinement process cannot generate high-quality images, if the initial images are badly generated.
Second, each word in an input sentence has a different level of information depicting the image content.
Current models utilize the same word representations in different image refinement processes, which makes the refinement process ineffective.
The image information should be taken into account to determine the importance of every word for refinement.

In this paper, we introduce a novel Dynamic Memory Generative Adversarial Network (DM-GAN) to address the aforementioned issues.
For the first issue, we propose to add a memory mechanism to cope with badly-generated initial images.
Recent work \cite{Anonymous:9_kqi-15} has shown the memory network's ability to encode knowledge sources. 
Inspired by this work, we propose to add the key-value memory structure \cite{Miller:2016uc} to the GAN framework. 
The fuzzy image features of initial images are treated as queries to read features from the memory module. 
The reads of the memory are used to refine the initial images.
To solve the second issue, we introduce a memory writing gate to dynamically select the words that are relevant to the generated image.
This makes our generated image well conditioned on the text description.
Therefore, the memory component is written and read dynamically at each image refinement process according to the initial image and text information.
In addition, instead of directly concatenating image and memory, a response gate is used to adaptively receive information from image and memory.

We conducted experiments to evaluate the DM-GAN model on the Caltech-UCSD Birds 200 (CUB) dataset and the Microsoft Common Objects in Context (COCO) dataset.
The quality of generated images is measured using the Inception Score (IS), the Fr\'{e}chet Inception Distance (FID) and the R-precision.
The experiments illustrate that our DM-GAN model outperforms the previous text-to-image synthesis methods, quantitatively and qualitatively.
Our model improves the IS from 4.36 to 4.75 and decreases the FID from 23.98 to 16.09 on the CUB dataset.
The R-precision is improved by 4.49\% and 3.09\% on the above two datasets.
The qualitative evaluation proves that our model generates more photo-realistic images.

This paper makes the following key contributions:
\begin{itemize}
    \item We propose a novel GAN model combined with a dynamic memory component to generate high-quality images even if the initial image is not well generated.
    \item We introduce a memory writing gate that is capable of selecting relevant word according to the initial image.
    \item A response gate is proposed to adaptively fuse information from image and memory.
    \item The experimental results demonstrate that the DM-GAN outperforms the state-of-the-art approaches.
\end{itemize}
\section{Related Work}
\subsection{Generative Adversarial Networks.}
With the recent successes of Variational Autoencoders (VAEs) \cite{kingma2013auto} and GANs \cite{Goodfellow:2014td}, a large number of methods have been proposed to handle generation \cite{Mirza:2014wi,Odena:2017vc,Xu:2017wg,cao2018adversarial} and domain adaptation task \cite{tzeng2017adversarial, zhu2019simreal}.
Recently, generating images based on the text descriptions gains interest in the research community nowadays.

\textbf{Single-stage.} The text-to-image synthesis problem is decomposed by Reed \etal \cite{Reed:2016ur} into two sub-problems:
first, the joint embedding is learned to capture the relations between natural language and real-world images; second, a deep convolutional generative adversarial network \cite{Radford:wf} is trained to synthesize a compelling image.
Dong \etal \cite{dong2017semantic} adopted the pair-wise ranking loss \cite{DBLP:journals/corr/KirosSZ14} to project both images and natural languages into a joint embedding space.
Since previous generative models failed to add the location information, Reed \etal proposed GAWWN \cite{reed2016learning} to encode localization constraints.
To diversify the generated images, the discriminator of TAC-GAN \cite{dash2017tac} not only distinguishes real images from synthetic images, but also classifies synthetic images into true classes.
Similar to TAC-GAN, PPGN \cite{nguyen2017plug} includes a conditional network to synthesize images conditioned on a caption.

\textbf{Multi-stage.} StackGAN \cite{Zhang:2017bw} and StackGAN++ \cite{zhang2017stackgan++} generate photo-realistic high-resolution images with two stages.
Yuan \etal \cite{yuan2018text} employed symmetrical distillation networks to minimize the multi-level difference between real and synthetic images.
DA-GAN \cite{ma2018gan} translates each word into a sub-region of an image.
Our method considers the interaction between each word and the whole generated image. 
Conditioning on the global sentence vector may result in low-quality images, AttnGAN \cite{Xu:2017wg} refines the images to high-resolution ones by leveraging the attention mechanism.
Each word in an input sentence has a different level of information depicting the image content.
However, AttnGAN takes all the words equally, it employs an attention module to use the same word representation. 
Our proposed memory module is able to uncover such difference for image generation, as it dynamically selects the important word information based on the initial image content. 


\begin{figure*}[!ht]
    \centering
    \includegraphics[width=1.0\textwidth]{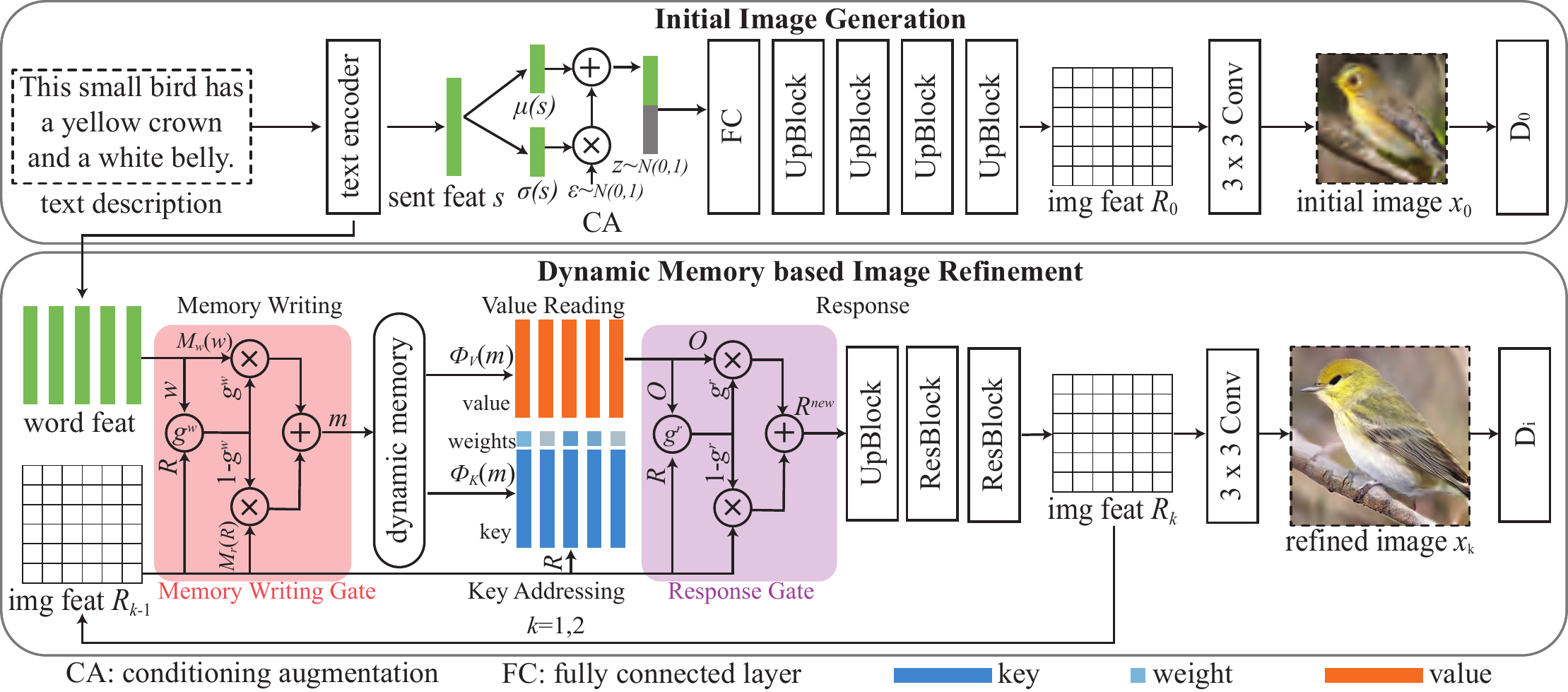}
    \caption{The DM-GAN architecture for text-to-image synthesis. Our DM-GAN model first generates an initial image, and then refines the initial image to generate a high-quality one. 
    }
    \label{fig:architecture}
\end{figure*}

\subsection{Memory Networks.}
Recently, memory network \cite{gulcehre2018dynamic,Anonymous:9_kqi-15} provides a new architecture to reason answers from memories more effectively using explicit storage and a notion of attention.
Memory network first writes information into an external memory and then reads contents from memory slots according to a relevance probability.
Weston \etal \cite{Anonymous:9_kqi-15} introduced the memory network to produce the output by searching supporting memories one by one.
End-to-end memory network \cite{Sukhbaatar:2015ww} is a continues form of memory network, where each memory slot is weighted according to the inner product between the memory and the query.
To understand the unstructured documents, the Key-Value Memory Network (KV-MemNN) \cite{Miller:2016uc} performs reasoning by utilizing different encodings for key memory and value memory. 
The key memory is used to infer the weight of the corresponding value memory when predicting the final answer.
Inspired by the recent success of the memory network, we introduce DM-GAN, a novel network architecture to generate high-quality images via nontrivial transforms between key and value memories.

\section{DM-GAN}
As shown in Figure \ref{fig:architecture}, the architecture of our DM-GAN model is composed of two stages: \textit{initial image generation} and \textit{dynamic memory based image refinement}.

At the \textit{initial image generation} stage, firstly, the input text description is transformed into some internal representation (a sentence feature $s$ and several word features $W$) by a text encoder.
Then, a deep conventional generator predicts an initial image $x_0$ with a rough shape and few details according to the sentence feature and a random noise vector $z$: $x_0, R_0 = G_0(z,s)$, where $R_{0}$ is the image feature.
The noise vector is sampled from a normal distribution.

At the \textit{dynamic memory based image refinement} stage, more fine-grained visual contents are added to the fuzzy initial images to generate a photo-realistic image $x_i$: $x_i=G_i(R_{i-1},W)$, where $R_{i-1}$ is the image feature from the last stage.
The refinement stage can be repeated multiple times to retrieve more pertinent information and generate a high-resolution image with more fine-grained details.

The dynamic memory based image refinement stage consists of four components: \textit{Memory Writing}, \textit{Key Addressing}, \textit{Value Reading}, and \textit{Response} (Section \ref{sec:dynamic}).
The \textit{Memory Writing} operation stores the text information into a key-value structured memory for further retrieval.
Then, \textit{Key Addressing} and \textit{Value Reading} operations are employed to read features from the memory module to refine the visual features of the low-quality images.
At last, the \textit{Response} operation is adopted to control the fusion of the image features and the reads of the memory.
We propose a memory writing gate to highlight important word information according to the image content in memory writing step (Section \ref{sec:gate1}).
We also utilize a response gate to adaptively fuse the information read from the memory and the image features (Section \ref{sec:gate2}).

\subsection{Dynamic Memory}\label{sec:dynamic}
We start with the given input word representations $W$, image $x$ and image features $R_i$:
\begin{align}
W &=\{w_1,w_2,...,w_T\},w_i\in \mathbb{R}^{N_w},\\
R_{i} &=\{r_1,r_2,...,r_N\},r_i\in \mathbb{R}^{N_r} ,
\end{align}
where $T$ is the number of words, $N_w$ is the dimension of word features, $N$ is the number of image pixels and image pixel feature is a $N_r$ dimensional vector.
We are intended to learn a model to refine the image using a more effective way to fuse text and image information via nontrivial transforms between key and value memory.
The refinement stage includes the following four steps.

\textbf{Memory Writing}: Encoding prior knowledge is an important part of the dynamic memory, which enables recovering high-quality images from text.
A naive way to write the memory is considering only partial text information.
\begin{equation}\label{eq:memorywriting}
    m_i = M(w_i), m_i \in \mathbb{R}^{N_m}
\end{equation}
where $M(\cdot)$ denotes the 1$\times$1 convolution operation which embeds word features into the memory feature space with $N_m$ dimensions.

\textbf{Key Addressing}: 
In this step, we retrieve relevant memories using key memory.
We compute a weight of each memory slot as a similarity probability between a memory slot $m_i$ and an image feature $r_j$:
\begin{equation}\label{eq:keyaddressing}
    \alpha_{i,j} = \frac{exp(\phi_K(m_i)^T r_j)}{\sum\limits_{l=1}^{T} exp(\phi_K(m_l)^T r_j)},
\end{equation}
where $\alpha_{i,j}$ is the similarity probability between the $i$-th memory and the $j$-th image feature and $\phi_K()$ is the key memory access process which maps memory features into dimension $N_r$. $\phi_K()$ is implemented as a 1$\times$1 convolution.
    
\textbf{Value Reading}:
The output memory representation is defined as the weighted summation of value memories according to the similarity probability:
\begin{equation}
    o_j = \sum_{i=1}^{T} \alpha_{i,j}\phi_V(m_i),
\end{equation}
where $\phi_V()$ is the value memory access process which maps memory features into dimension $N_r$. $\phi_V()$ is implemented as a 1$\times$1 convolution.
    
\textbf{Response}:
After receiving the output memory, we combine the current image and the output representation to provide a new image feature.
A naive approach will be simply concatenating the image features and the output representation.
The new image features are obtained by:
\begin{equation}\label{eq:response}
    r_i^{new} = [o_i,r_i],
\end{equation}
where $[\cdot,\cdot]$ denotes concatenation operation.
Then, we are able to utilize an upsampling block and several residual blocks \cite{He2015} to upscale the new image features into a high-resolution image.
The upsampling block consists of a nearest neighbor upsampling layer and a 3$\times$3 convolution.
Finally, the refined image $x$ is obtained from the new image features using a 3$\times$3 convolution.

\subsection{Gated Memory Writing}\label{sec:gate1}
Instead of considering only partial text information using Eq.\ref{eq:memorywriting}, the memory writing gate allows the DM-GAN model to select the relevant word to refine the initial images.
The memory writing gate $g_i^w$ combines image features $R_i$ from the last stage with word features $W$ to calculate the importance of a word:
\begin{equation}\label{eq:gatedmemorywriting}
    g_i^w(R,w_i) = \sigma(A*w_i + B*\frac{1}{N}\sum\limits_{i=1}^{N}r_i),
\end{equation}
where $\sigma$ is the sigmoid function, $A$ is a $1 \times N_w$ matrix, and $B$ is a $1 \times N_r$ matrix.
Then, the memory slot $m_i\in R^{N_m}$ is written by combining the image and word features.
\begin{equation}\label{eq:mw2}
    m_i = M_w(w_i)*g_i^w + M_r(\frac{1}{N}\sum\limits_{i=1}^{N}r_i)*(1-g_i^w),
\end{equation}
where $M_w(\cdot)$ and $M_r(\cdot)$ denote the 1x1 convolution operation.
$M_w(\cdot)$ and $M_r(\cdot)$ embed image and word features into the same feature space with $N_m$ dimensions.
    
\subsection{Gated Response}\label{sec:gate2}
We utilize the adaptive gating mechanism to dynamically control the information flow and update image features:
\begin{equation}\label{eq:gatedresponse}
    \begin{split}
    g^r_i &= \sigma(W[o_i,r_i]+b),\\
    r^{new}_i& = o_i * g^r_i + r_i * (1 - g^r_i),
    \end{split}
\end{equation}
where $g^r_i$ is the response gate for information fusion, $\sigma$ is the sigmoid function, $W$ and $b$ are the parameter matrix and bias term.

\subsection{Objective Function}
The objective function of the generator network is defined as:
\begin{equation}
    L =  \sum_iL_{G_i} +  \lambda_1 L_{CA} + \lambda_2 L_{DAMSM},
\end{equation}
in which $\lambda_1$ and $\lambda_2$ are the corresponding weights of conditioning augmentation loss and DAMSM loss. $G_0$ denotes the generator of the initial generation stage. $G_i$ denotes the generator of the $i$-th iteration of the image refinement stage.

\textbf{Adversarial Loss}: The adversarial loss for $G_i$ is defined as follows:
\begin{equation}
L_{G_i}\!=\! -\frac{1}{2}[\mathbb{E}_{x \sim p_{G_i}}logD_i(x)+\mathbb{E}_{x\sim p_{G_i}}logD_i(x,s)],
\end{equation}
where the first term is the unconditional loss which makes the generated image real as much as possible and the second term is the conditional loss which makes the image match the input sentence.
Alternatively, the adversarial loss for each discriminator $D_i$ is defined as:
\begin{equation}
    \begin{split}
     L_{D_i}\!&=\! \underbrace{-\frac{1}{2}[\mathbb{E}_{x\sim p_{data}}log D_i (x)\!+\!\mathbb{E}_{x\sim p_{G_i}}log(1\!-\!D_i (x))}_{\text{ unconditional loss}},\\
    & \underbrace{+\mathbb{E}_{x\sim p_{data}}log D_i (x,s) \!+\!\mathbb{E}_{x\sim p_{G_i}}log(1\!-\!D_i (x,s))]}_{\text{ conditional loss}},
    \end{split}
\end{equation}
where the unconditional loss is designed to distinguish the generated image from real images and the conditional loss determines whether the image and the input sentence match.

\textbf{Conditioning Augmentation Loss}: The Conditioning Augmentation (CA) technique \cite{Zhang:2017bw} is proposed to augment training data and avoid overfitting by resampling the input sentence vector from an independent Gaussian distribution.
Thus, the CA loss is defined as the Kullback-Leibler divergence between the standard Gaussian distribution and the Gaussian distribution of training data.
\begin{equation}
    L_{CA} = D_{KL}(\mathcal{N}(\mu(s),\Sigma(s))||\mathcal{N}(0,I)),
\end{equation}
where $\mu(s)$ and $\Sigma(s)$ are mean and diagonal covariance matrix of the sentence feature. $\mu(s)$ and $\Sigma(s)$ are computed by fully connected layers.

\textbf{DAMSM Loss}: We utilize the DAMSM loss \cite{Xu:2017wg} to measure the matching degree between images and text descriptions. The DAMSM loss makes generated images better conditioned on text descriptions.

\subsection{Implementation Details} 
For text embedding, we employ a pre-trained bidirectional LSTM text encoder by Xu \etal \cite{Xu:2017wg} and fix their parameters during training.
Each word feature corresponds to the hidden states of two directions.
The sentence feature is generated by concatenating the last hidden states of two directions.
The initial image generation stage first synthesizes images with 64x64 resolution.
Then, the dynamic memory based image refinement stage refines images to 128x128 and 256x256 resolution.
We only repeat the refinement process with dynamic memory module two times due to GPU memory limitation.
Introducing dynamic memory to low-resolution images (\ie 16x16, 32x32) can not further improve the performance.  
Because low-resolution images are not well generated and their features are more like random vectors.
For all discriminator networks, we apply spectral normalization \cite{Miyato:2018wa} after every convolution to avoid unusual gradients to improve text-to-image synthesis performance.
By default, we set $N_w=256$, $N_r=64$ and $N_m=128$ to be the dimension of text, image and memory feature vectors respectively.
We set the hyperparameter $\lambda_1=1$ and $\lambda_2=5$ for the CUB dataset and $\lambda_1=1$ and $\lambda_2=50$ for the COCO dataset.
All networks are trained using ADAM optimizer \cite{kingma2014adam} with batch size 10, $\beta_1=0.5$ and $\beta_2=0.999$.
The learning rate is set to be 0.0002.
We train the DM-GAN model with 600 epochs on the CUB dataset and 120 epochs on the COCO dataset.
\begin{table*}[!ht]\setlength{\tabcolsep}{8pt}
\centering
\begin{tabular}{lcccccc}
\toprule
Dataset        & GAN-INT-CLS \cite{Reed:2016ur}  & GAWWN \cite{reed2016learning}        & StackGAN \cite{Zhang:2017bw}      & PPGN \cite{nguyen2017plug}         & AttnGAN \cite{Xu:2017wg}       & DM-GAN     \\ \midrule
CUB      & 2.88$\pm$0.04 & 3.62$\pm$0.07 & 3.70$\pm$0.04 & (-)           & 4.36$\pm$0.03  & \textbf{4.75$\pm$0.07}  \\ \midrule
COCO     & 7.88$\pm$0.07 & (-)           & 8.45$\pm$0.03 & 9.58$\pm$0.21 & 25.89$\pm$0.47 & \textbf{30.49$\pm$0.57}\\ 
\bottomrule
\end{tabular}
\caption{The inception scores (higher is better) of GAN-INT-CLS \cite{Reed:2016ur}, GAWWN \cite{reed2016learning}, StackGAN \cite{Zhang:2017bw}, PPGN \cite{nguyen2017plug}, AttnGAN \cite{Xu:2017wg} and our DM-GAN on the CUB and COCO datasets. The best results are in bold.}
\label{table:IS}
\end{table*}

\begin{table}[!t]\setlength{\tabcolsep}{6.5pt}
\begin{tabular}{llcc}
\toprule
Dataset & Metric & AttnGAN & DM-GAN \\ \midrule
CUB & FID$\downarrow$ & 23.98 & \textbf{16.09} \\
 & R-precision$\uparrow$ & 67.82$\pm$4.43 & \textbf{72.31$\pm$0.91}\\ \midrule
COCO & FID$\downarrow$ & 35.49 & \textbf{32.64} \\
 & R-precision$\uparrow$ & 85.47$\pm$3.69 & \textbf{88.56$\pm$0.28}\\
\bottomrule
\end{tabular}
\caption{Performance of FID and R-precision for AttnGAN \cite{Xu:2017wg} and our DM-GAN on the CUB and COCO datasets. The FID of AttnGAN is calculated from officially released weights. Lower is better for FID and higher is better for R-precision.}
\label{table:FID}
\end{table}

\section{Experiments}
In this section, we evaluate the DM-GAN model quantitatively and qualitatively.
We implemented the DM-GAN model using the open-source Python library PyTorch \cite{paszke2017automatic}. 

\textbf{Datasets.} To demonstrate the capability of our proposed method for text-to-image synthesis, we conducted experiments on the CUB \cite{wah2011caltech} and the COCO \cite{lin2014microsoft} datasets.
The CUB dataset contains 200 bird categories with 11,788 images, where 150 categories with 8,855 images are employed for training while the remaining 50 categories with 2,933 images for testing. There are ten captions for each image in CUB dataset.
The COCO dataset includes a training set with 80k images and a test set with 40k images. 
Each image in the COCO dataset has five text descriptions.

\textbf{Evaluation Metric.} We quantify the performance of the DM-GAN in terms of Inception Score (IS), Fr\'{e}chet Inception Distance (FID), and R-precision. Each model generated 30,000 images conditioning on the text descriptions from the unseen test set for evaluation.

The IS \cite{Salimans:2016wg} uses a pre-trained Inception v3 network \cite{szegedy2016rethinking} to compute the KL-divergence between the conditional class distribution and the marginal class distribution.
A large IS means that the generated model outputs a high diversity of images for all classes and each image clearly belongs to a specific class.

The FID \cite{heusel2017gans} computes the Fr\'{e}chet distance between synthetic and real-world images based on the extracted features from a pre-trained Inception v3 network. A lower FID implies a closer distance between generated image distribution and real-world image distribution.

Following Xu \etal \cite{Xu:2017wg}, we use the R-precision to evaluate whether a generated image is well conditioned on the given text description.
The R-precision is measured by retrieving relevant text given an image query.
We compute the cosine distance between a global image vector and 100 candidate sentence vectors. The candidate text descriptions include R ground truth and 100-R randomly selected mismatching descriptions. For each query, if r results in the top R ranked retrieval descriptions are relevant, then the R-precision is r/R.
In practice, we compute the R-precision with R=1.
We divide the generated images into ten folds for retrieval and then take the mean and standard deviation of the resulting scores.

\subsection{Text-to-Image Quality}
We compare our DM-GAN model with the state-of-the-art models on the CUB and COCO test datasets.
The performance results are reported in Table \ref{table:IS} and \ref{table:FID}.

As shown in Table \ref{table:IS}, our DM-GAN model achieves 4.75 IS on the CUB dataset, which outperforms other methods by a large margin.
Compared with AttnGAN, DM-GAN improves the IS from 4.36 to 4.75 on the CUB dataset (8.94\% improvement) and from 25.89 to 30.49 on the COCO dataset (17.77\% improvement).
The experimental results indicate that our DM-GAN model generates images with higher quality than other approaches.

Table \ref{table:FID} compares the performance between AttnGAN and DM-GAN with respect to the FID on the CUB and COCO datasets. 
We measure the FID of AttnGAN from the officially pre-trained model.
Our DM-GAN decreases the FID from 23.98 to 16.09 on the CUB dataset and from 35.49 to 32.64 on the COCO dataset, which demonstrates that DM-GAN learns a better data distribution. 

As shown in Table \ref{table:FID}, the DM-GAN improves the R-precision by 4.49\% on the CUB dataset and 3.09\% on the COCO dataset. Higher R-precision indicates that the generated images by the DM-GAN are better conditioned on the given text description, which further demonstrates the effectiveness of the employed dynamic memory.

In summary, the experimental results indicate that our DM-GAN is superior to the state-of-the-art models.

\begin{figure*}[!ht]
    \centering
    \includegraphics[width=0.99\textwidth]{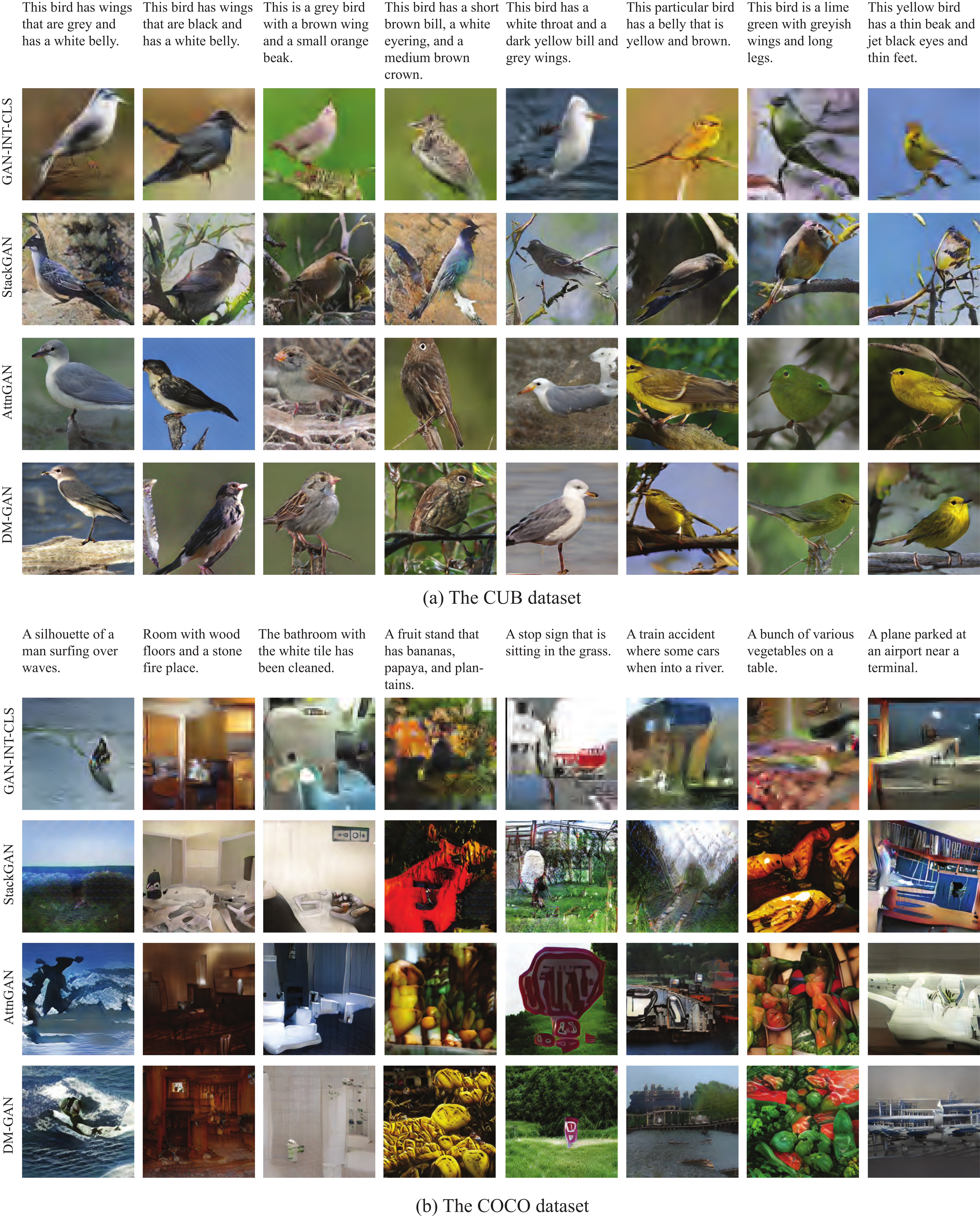}
    \caption{Example results for text-to-image synthesis by DM-GAN and AttnGAN. (a) Generated bird images by conditioning on text from CUB test set. (b) Generated images by conditioning on text from COCO test set.}
    \label{fig:visualization}
\end{figure*}
\begin{figure*}[!ht]
    \centering
    \includegraphics[width=0.99\textwidth]{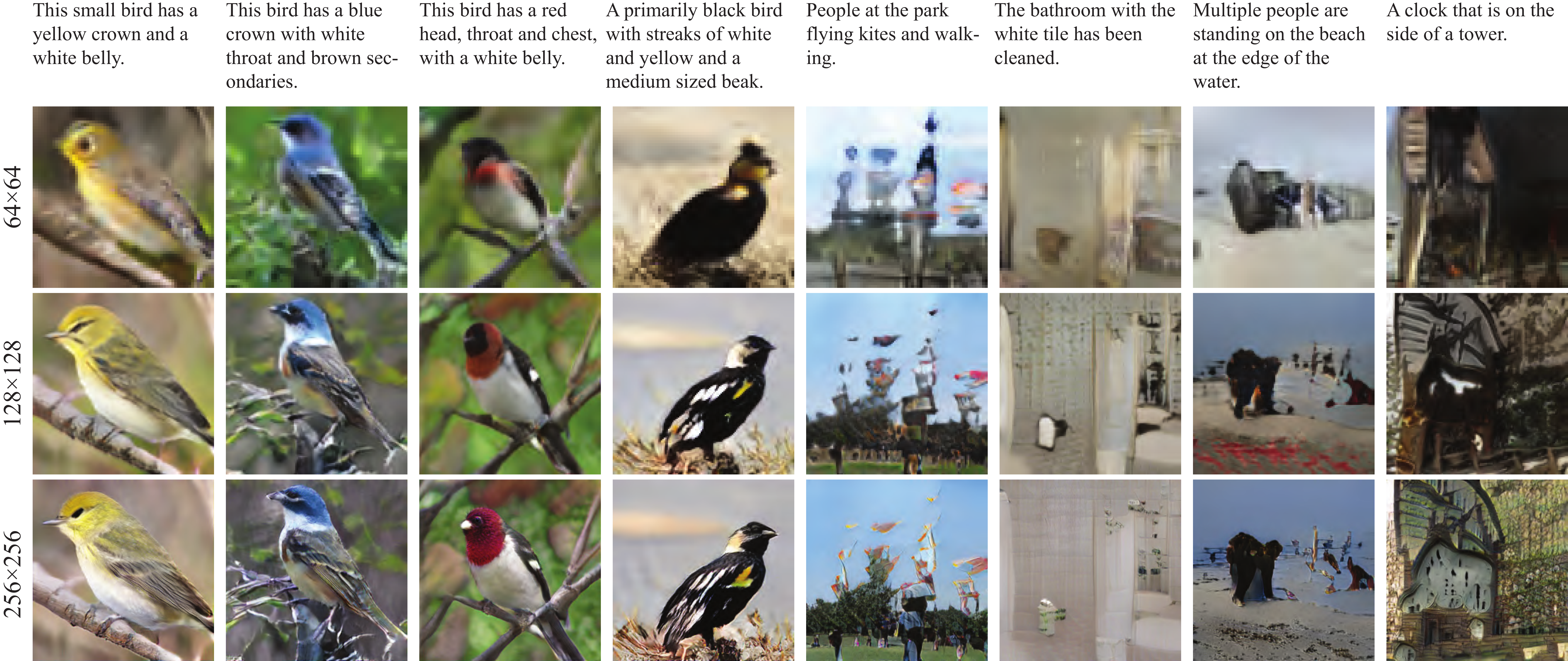}
    \caption{The results of different stages of our DM-GAN model, including the initial images, the images after one refinement process and the images after two refinement processes.
    }
    \label{fig:multistage}
\end{figure*}

\subsection{Visual Quality}
For qualitative evaluation, Figure \ref{fig:visualization} shows text-to-image synthesis examples generated by our DM-GAN and the state-of-the-art models.
In general, our DM-GAN approach generates images with more vivid details as well as more clear backgrounds in most cases, comparing to the AttnGAN \cite{Xu:2017wg}, GAN-INT-CLS \cite{Reed:2016ur} and StackGAN \cite{Zhang:2017bw}, because it employs a dynamic memory model using varied weighted word information to improve image quality.

Our DM-GAN method has the capacity to better understand the logic of the text description and present a more clear structure of the images.
Observing the samples generated on the CUB dataset in Figure \ref{fig:visualization}(a), with a single character, although DM-GAN and AttnGAN both perform well in accurately capture and present the character's feature, our DM-GAN model better highlights the main subject of the image, the bird, differentiating from its background.
It demonstrates that, with the dynamic memory module, our DM-GAN model is able to bridge the gap between visual contents and natural languages.
In terms of multi-subjects-image generation, for example, the COCO dataset in Figure \ref{fig:visualization}(b), it is more challenging to generate photo-realistic images when the text description is more complicated and contains more than one subject.
DM-GAN precisely captures the major scene based on the most important subject and arrange the rest descriptive contents logically, which improves the global structure of the image. For instance, DM-GAN is the only successful method clearly identifies the bathroom with required components in the column 3 in Figure \ref{fig:visualization}(b).
The visual results show that our DM-GAN is more effective to capture important subjects using a memory writing gate to dynamically select important words.

Figure \ref{fig:multistage} indicates that our DM-GAN model is able to refine badly initialized images and generate more photo-realistic high-resolution images. So the image quality is obviously well-improved, with clear backgrounds and convincing details.
In most cases, the initial stage generates a blurry image with rough shape and color, so that the background is fine-tuned to be more realistic with fine-grained textures, while the refined image will be better conditioned on the input text and provide more photo-realistic high-resolution images.
In the fourth column of Figure \ref{fig:multistage}, no white streaks can be found on the bird's body from the initial image with 64$\times$64 resolution.
The refinement process helps to encode "white streaks" information from text description and add back missing features based on the text description and image content.
In order word, our DM-GAN model is able to refine the image to match the input text description.

\begin{figure}
    \centering
    \includegraphics[width=\columnwidth]{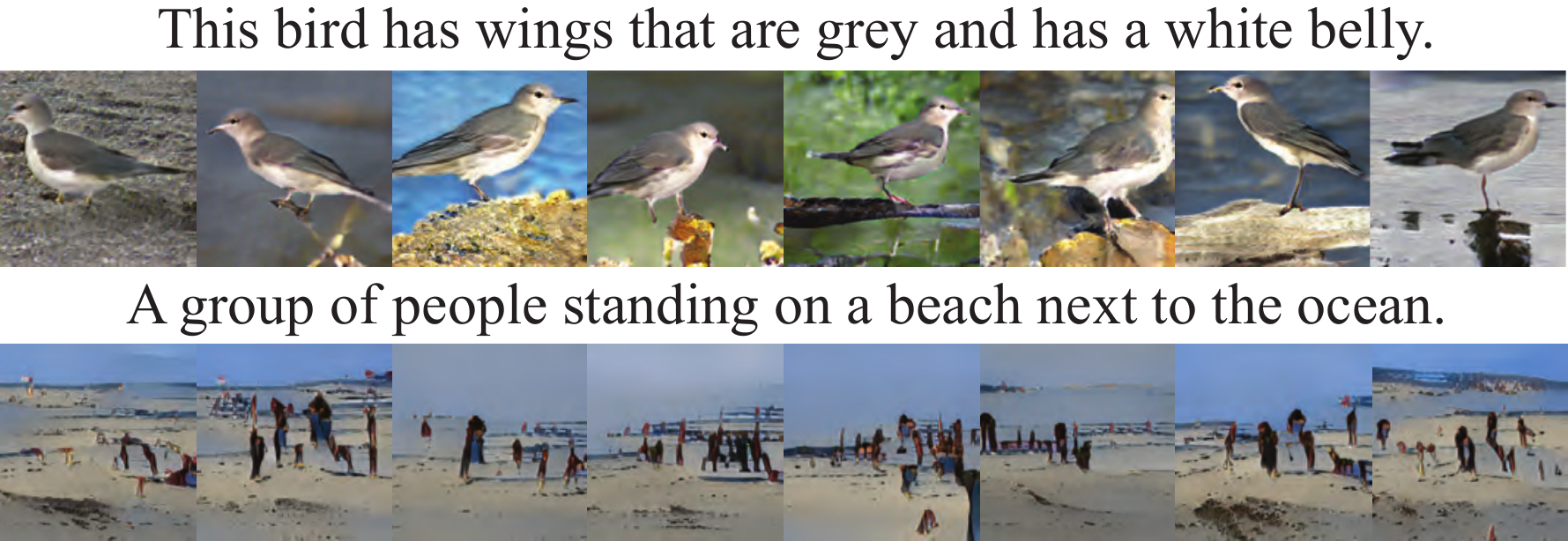}
    \caption{Generated images using the same text description.}
    \label{fig:diversity}
\end{figure}

To evaluate the diversity of our DM-GAN model, we generate several images using the same text description, and multiple noise vectors.
Figure \ref{fig:diversity} shows text descriptions and synthetic images with different shapes and backgrounds.
Images are similar but not identical to each other, which means our DM-GAN generates images with high diversity.

\begin{table}[t]\setlength{\tabcolsep}{8pt}
\centering
\begin{tabular}{lccc}
\toprule
Architecture      & IS$\uparrow$ & FID$\downarrow$ & R-Precision$\uparrow$\\\midrule
baseline         &  4.51$\pm$0.04  &  23.32   & 68.60$\pm$0.73\\
+M       &  4.57$\pm$0.05  &  21.41  & 70.66$\pm$0.69\\
+M+WG    &  4.65$\pm$0.05  &  20.83  & 71.40$\pm$0.64\\
+M+WG+RG &  \textbf{4.75$\pm$0.07}  &  \textbf{16.09} & \textbf{72.31$\pm$0.91}\\
\bottomrule
\end{tabular}
\caption{The performance of different architectures of our DM-GAN on the CUB  datasets. M, WG and RG denote dynamic memory, memory writing gate and response gate respectively.}
\label{tab:ca}
\end{table}

\subsection{Ablation Study}
In order to verify the effectiveness of our proposed components, we evaluate the DM-GAN architecture and its variants on the CUB dataset.
The control components between architectures include the key-value memory (M), the writing gate (WG) and the response gate (RG).
We define a baseline model which removes M, WG and RG from DM-GAN.
The memory is written according to partial text information (Eq.\ref{eq:memorywriting}).
The response operation simply concatenates the image features and the memory output (Eq.\ref{eq:response}).
The performance of the DM-GAN architecture and its variants is reported in Table \ref{tab:ca}.
Our baseline model produces slightly better performance than AttnGAN.
By integrating these components, our model can achieve further improvement which demonstrates the effectiveness of every component.

\begin{figure}[!h]
    \centering
    \includegraphics[width=0.99\columnwidth]{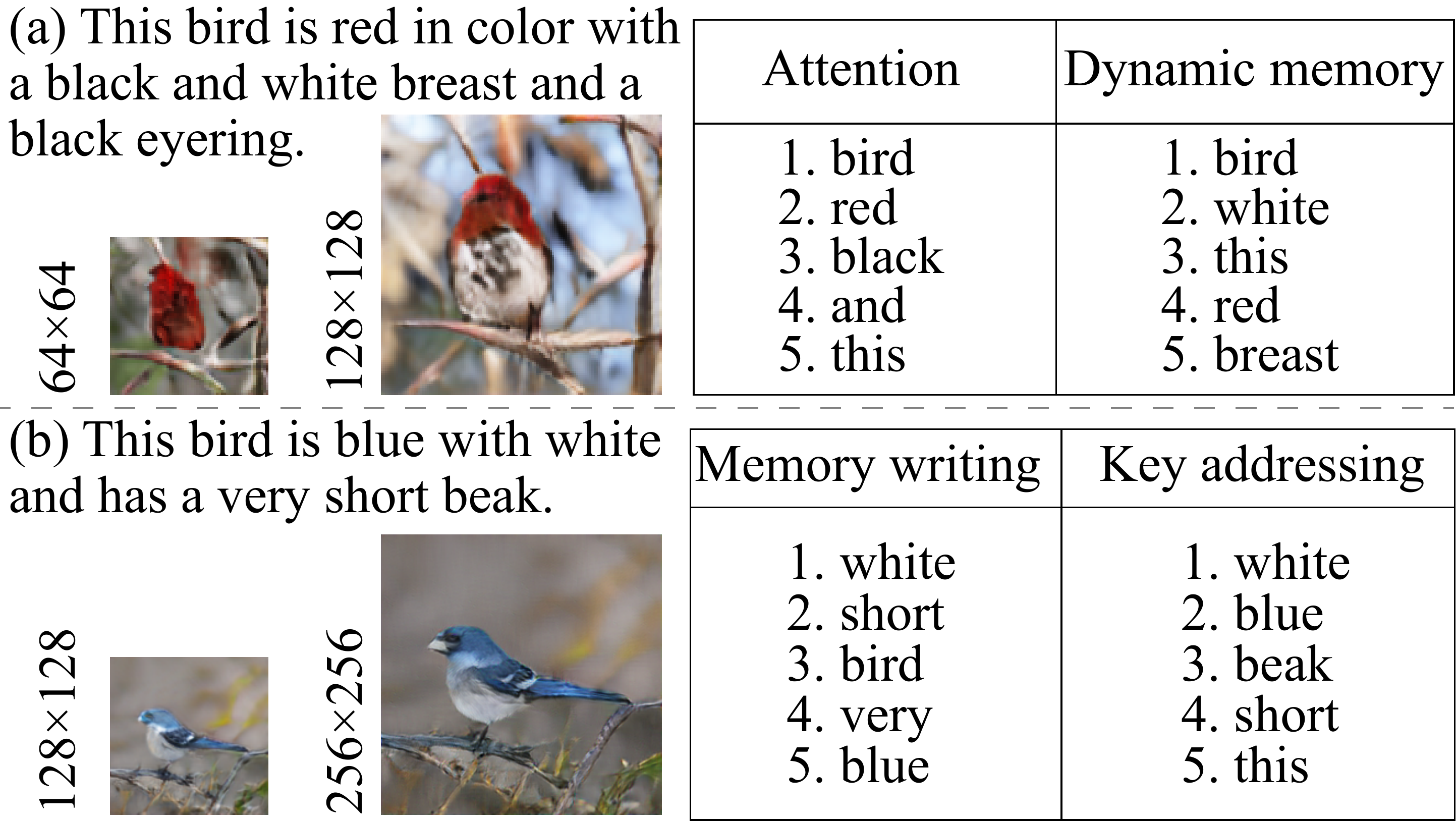}
    \caption{(a) Comparison between the top 5 relevant words selected by attention module and dynamic memory module. (b) The top 5 relevant words selected by memory writing step and key addressing step.}
    \label{fig:example1}
\end{figure}

Further, we visualize the most relevant words selected by the AttnGAN \cite{Xu:2017wg} and our DM-GAN.
We notice that the attention mechanism cannot accurately select relevant words when the initial images are not well generated.
We propose the dynamic memory module to select the most relevant words based on the global image feature.
As Fig. \ref{fig:example1} (a) shows, although a bird with incorrect red breast is generated, dynamic memory module selects the word, \emph{i.e.}, "white" to correct the image.
The DM-GAN selects and combines word information with image features in two steps (see Fig. \ref{fig:example1} (b)). 
The gated memory writing step first roughly selects words relevant to the image and writes them into the memory. Then the key addressing step further reads more relevant words from the memory.

\section{Conclusions}
In this paper, we have proposed a new architecture called DM-GAN for text-to-image synthesis task.
We employ a dynamic memory component to refine the initial generated image, a memory writing gate to highlight important text information and a repose gate to fuse image and memory representation.
Experiment results on two real-world datasets show that DM-GAN outperforms the state-of-the-art by both qualitative and quantitative measures.
Our DA-GAN refines initial images with wrong color and rough shapes. However, the final results still rely heavily on the layout of multi-subjects in initial images.
In the future, we will try to design a more powerful model to generate initial images with better organizations. 

\section*{Acknowledgment}
This research has been supported by National Key Research and Development Program (2018YFB0904503) and National Natural Science Foundation of China (U1866602, 61772456).




{\small
\bibliographystyle{ieee}
\bibliography{reference}

\begin{thebibliography}{10}\itemsep=-1pt

\bibitem{cao2018adversarial}
Jiezhang Cao, Yong Guo, Qingyao Wu, Chunhua Shen, and Mingkui Tan.
\newblock Adversarial learning with local coordinate coding.
\newblock {\em ICML}, 2018.

\bibitem{dash2017tac}
Ayushman Dash, John Cristian~Borges Gamboa, Sheraz Ahmed, Marcus Liwicki, and
  Muhammad~Zeshan Afzal.
\newblock Tac-gan-text conditioned auxiliary classifier generative adversarial
  network.
\newblock {\em arXiv preprint arXiv:1703.06412}, 2017.

\bibitem{dong2017semantic}
Hao Dong, Simiao Yu, Chao Wu, and Yike Guo.
\newblock Semantic image synthesis via adversarial learning.
\newblock In {\em Proceedings of the IEEE ICCV}, pages 5706--5714, 2017.

\bibitem{Goodfellow:2014td}
Ian Goodfellow, Jean Pouget-Abadie, Mehdi Mirza, Bing Xu, David Warde-Farley,
  Sherjil Ozair, Aaron Courville, and Yoshua Bengio.
\newblock {Generative Adversarial Nets}.
\newblock In {\em NIPS}, pages 2672--2680, 2014.

\bibitem{gulcehre2018dynamic}
Caglar Gulcehre, Sarath Chandar, Kyunghyun Cho, and Yoshua Bengio.
\newblock Dynamic neural turing machine with continuous and discrete addressing
  schemes.
\newblock {\em Neural computation}, 30(4):857--884, 2018.

\bibitem{He2015}
Kaiming He, Xiangyu Zhang, Shaoqing Ren, and Jian Sun.
\newblock Deep residual learning for image recognition.
\newblock {\em arXiv preprint arXiv:1512.03385}, 2015.

\bibitem{heusel2017gans}
Martin Heusel, Hubert Ramsauer, Thomas Unterthiner, Bernhard Nessler, and Sepp
  Hochreiter.
\newblock Gans trained by a two time-scale update rule converge to a local nash
  equilibrium.
\newblock In {\em NIPS}, pages 6626--6637, 2017.

\bibitem{kingma2014adam}
D Kinga and J~Ba Adam.
\newblock A method for stochastic optimization.
\newblock In {\em ICLR}, volume~5, 2015.

\bibitem{kingma2013auto}
Diederik~P Kingma and Max Welling.
\newblock Auto-encoding variational bayes.
\newblock {\em arXiv preprint arXiv:1312.6114}, 2013.

\bibitem{DBLP:journals/corr/KirosSZ14}
Ryan Kiros, Ruslan Salakhutdinov, and Richard~S. Zemel.
\newblock Unifying visual-semantic embeddings with multimodal neural language
  models.
\newblock {\em CoRR}, abs/1411.2539, 2014.

\bibitem{lin2014microsoft}
Tsung-Yi Lin, Michael Maire, Serge Belongie, James Hays, Pietro Perona, Deva
  Ramanan, Piotr Doll{\'a}r, and C~Lawrence Zitnick.
\newblock Microsoft coco: Common objects in context.
\newblock In {\em ECCV}, pages 740--755. Springer, 2014.

\bibitem{ma2018gan}
Shuang Ma, Jianlong Fu, Chang Wen~Chen, and Tao Mei.
\newblock Da-gan: Instance-level image translation by deep attention generative
  adversarial networks.
\newblock In {\em CVPR}, pages 5657--5666, 2018.

\bibitem{Miller:2016uc}
Alexander Miller, Adam Fisch, Jesse Dodge, Amir-Hossein Karimi, Antoine Bordes,
  and Jason Weston.
\newblock Key-value memory networks for directly reading documents.
\newblock In {\em ACL}, pages 1400--1409, 2016.

\bibitem{Mirza:2014wi}
Mehdi Mirza and Simon Osindero.
\newblock {Conditional Generative Adversarial Nets.}
\newblock {\em CoRR}, 2014.

\bibitem{Miyato:2018wa}
Takeru Miyato, Toshiki Kataoka, Masanori Koyama, and Yuichi Yoshida.
\newblock Spectral normalization for generative adversarial networks.
\newblock In {\em ICLR}, 2018.

\bibitem{nguyen2017plug}
Anh Nguyen, Jeff Clune, Yoshua Bengio, Alexey Dosovitskiy, and Jason Yosinski.
\newblock Plug \& play generative networks: Conditional iterative generation of
  images in latent space.
\newblock In {\em CVPR}, pages 4467--4477, 2017.

\bibitem{Odena:2017vc}
Augustus Odena, Christopher Olah, and Jonathon Shlens.
\newblock Conditional image synthesis with auxiliary classifier gans.
\newblock In {\em ICML}, pages 2642--2651, 2017.

\bibitem{paszke2017automatic}
Adam Paszke, Sam Gross, Soumith Chintala, Gregory Chanan, Edward Yang, Zachary
  DeVito, Zeming Lin, Alban Desmaison, Luca Antiga, and Adam Lerer.
\newblock Automatic differentiation in pytorch.
\newblock In {\em NIPS-W}, 2017.

\bibitem{Radford:wf}
Alec Radford, Luke Metz, and Soumith Chintala.
\newblock Unsupervised representation learning with deep convolutional
  generative adversarial networks.
\newblock {\em arXiv preprint arXiv:1511.06434}, 2015.

\bibitem{Reed:2016ur}
Scott Reed, Zeynep Akata, Xinchen Yan, Lajanugen Logeswaran, Bernt Schiele, and
  Honglak Lee.
\newblock Generative adversarial text to image synthesis.
\newblock In {\em ICML}, pages 1060--1069, 2016.

\bibitem{reed2016learning}
Scott~E Reed, Zeynep Akata, Santosh Mohan, Samuel Tenka, Bernt Schiele, and
  Honglak Lee.
\newblock Learning what and where to draw.
\newblock In {\em NIPS}, pages 217--225, 2016.

\bibitem{Salimans:2016wg}
Tim Salimans, Ian Goodfellow, Wojciech Zaremba, Vicki Cheung, Alec Radford, and
  Xi Chen.
\newblock Improved techniques for training gans.
\newblock In {\em NIPS}, pages 2234--2242, 2016.

\bibitem{Sukhbaatar:2015ww}
Sainbayar Sukhbaatar, Arthur Szlam, Jason Weston, and Rob Fergus.
\newblock End-to-end memory networks.
\newblock In {\em NIPS}, pages 2440--2448, 2015.

\bibitem{szegedy2016rethinking}
Christian Szegedy, Vincent Vanhoucke, Sergey Ioffe, Jon Shlens, and Zbigniew
  Wojna.
\newblock Rethinking the inception architecture for computer vision.
\newblock In {\em CVPR}, pages 2818--2826, 2016.

\bibitem{tzeng2017adversarial}
Eric Tzeng, Judy Hoffman, Kate Saenko, and Trevor Darrell.
\newblock Adversarial discriminative domain adaptation.
\newblock In {\em CVPR}, pages 7167--7176, 2017.

\bibitem{wah2011caltech}
Catherine Wah, Steve Branson, Peter Welinder, Pietro Perona, and Serge
  Belongie.
\newblock The caltech-ucsd birds-200-2011 dataset.
\newblock 2011.

\bibitem{Anonymous:9_kqi-15}
Jason Weston, Sumit Chopra, and Antoine Bordes.
\newblock {Memory Networks}.
\newblock In {\em ICLR}, 2015.

\bibitem{Xu:2017wg}
Tao Xu, Pengchuan Zhang, Qiuyuan Huang, Han Zhang, Zhe Gan, Xiaolei Huang, and
  Xiaodong He.
\newblock Attngan: Fine-grained text to image generation with attentional
  generative adversarial networks.
\newblock In {\em CVPR}, 2018.

\bibitem{yuan2018text}
Mingkuan Yuan and Yuxin Peng.
\newblock Text-to-image synthesis via symmetrical distillation networks.
\newblock {\em arXiv preprint arXiv:1808.06801}, 2018.

\bibitem{Zhang:2017bw}
Han Zhang, Tao Xu, and Hongsheng Li.
\newblock Stackgan: Text to photo-realistic image synthesis with stacked
  generative adversarial networks.
\newblock In {\em IEEE ICCV}, pages 5908--5916. IEEE, 2017.

\bibitem{zhang2017stackgan++}
H. Zhang, T. Xu, H. Li, S. Zhang, X. Wang, X. Huang, and D.~N. Metaxas.
\newblock Stackgan++: Realistic image synthesis with stacked generative
  adversarial networks.
\newblock {\em TPAMI}, 2018.

\bibitem{zhu2019simreal}
Fengda Zhu, Linchao Zhu, and Yi Yang.
\newblock Sim-real joint reinforcement transfer for 3d indoor navigation.
\newblock In {\em CVPR}, 2019.

\bibitem{Zhu_2017_CVPR}
Linchao Zhu, Zhongwen Xu, and Yi Yang.
\newblock Bidirectional multirate reconstruction for temporal modeling in
  videos.
\newblock In {\em CVPR}, July 2017.

\end{thebibliography}
}

\end{document}